\documentclass[pmlr,twocolumn,10pt]{jmlr} 
\usepackage{booktabs}
\usepackage{siunitx}

\usepackage[switch]{lineno}

\jmlrvolume{}
\jmlryear{}
\jmlrsubmitted{}
\jmlrpublished{}
\jmlrworkshop{}

 \title[Beyond validation loss]{Beyond validation loss: Clinically-tailored optimization metrics improve a model's clinical performance}

\author{%
    \Name{Charles B. Delahunt 1, 2} \Email{delahunt@uw.edu}\\
    \Name{Courosh Mehanian 1} \Email{courosh@uoregon.edu}\\
   \Name{Daniel Shea 1} \Email{shea.dan@gmail.com}\\
    \Name{Matthew P. Horning 1}  \Email{matthew.p.horning@gmail.com}\\ 
    \addr 1 formerly Global Health Labs, Bellevue, WA. \\
    2 University of Washington, Seattle, WA.
}

\begin{document}

\maketitle

\begin{abstract}
A key task in ML is to optimize models at various stages, e.g. by choosing hyperparameters or picking a stopping point.  
A traditional ML approach is to use validation loss, i.e. to apply the training loss function on a validation set to guide these optimizations.

However, ML for healthcare has a distinct goal from traditional ML: Models must perform well relative to specific clinical requirements, vs. relative to the loss function used for training.
These clinical requirements can be captured more precisely by tailored metrics.
Since many optimization tasks do not require the driving metric to be differentiable, they allow a wider range of options, including  the use of metrics tailored to be clinically-relevant.

In this paper we describe two controlled experiments which show how the use of clinically-tailored metrics yields superior model optimization compared to validation loss, in the sense of better performance on the clinical task.  

The use of clinically-relevant metrics for optimization entails some extra effort, to  define the metrics and to code them into the pipeline.
But it can yield models that better meet the central goal of ML for healthcare: strong performance in the clinic.

\end{abstract}

\paragraph*{Data and Code Availability}
The \textit{Loa loa} dataset is in preparation for open-access release \citep{kamgno_loaDataset}. 
The fetal ultrasound dataset \citep{pokaprakarn_blindSweep} 
is not yet available publicly.
Relevant Python code is included in \appendixref{apd:codeFragments}.

\paragraph*{Institutional Review Board (IRB)} ~\\ 
(1) CE N\textsuperscript{o} 0094/CRERSHC/2023 Yaound\'e, Cameroon (\textit{Loa loa} study). (2) UNC School of Medicine IRB\# 24-1415 (ultrasound study).



\section{Introduction}
\label{sec:intro}
Metrics are fundamental to optimizing machine learning (ML) algorithms, including tasks such as hyperparameter optimization and selection of stopping point for training. 
Certain metrics inherited from ML, e.g. cross-entropy loss, are engrained in ML convention, are straight-forward to implement thanks to well-curated libraries, and are often highly effective from an ML perspective. 
In addition, using the same loss function on train and validation sets is automatic in ML frameworks such as PyTorch.
As a result these metrics are embedded in the practice of ML as a standard way to optimize models, on the implicit assumption that they will lead to an optimization that performs well in the clinic. 
 
However, in ML for health care, performing well relative to the training loss function is not a model's actual goal.
Rather, the true goal of a model developed for a medical use case is to meet the specifications determined by the clinical needs of that use case.
These needs are characterized by metrics that often diverge significantly from standard ML metrics.
 
A clear example is object-level vs. patient-level metrics: 
An algorithm is often trained at the object-level - that is, the unit passed through the model is a patch of a histology slide, a suspected parasite, a thumbnail of an object of interest, etc, and algorithm performance is optimized using metrics at this level; in contrast, the clinician cares about a patient, whose sample under test contains many objects. 
There is necessarily a transform, usually non-linear, from object-level results to patient-level results. 
For example, the algorithm might evaluate several image patches from a histology slide, then patient disposition is based on some function of the patch results.
A model optimized for maximum performance at the object-level has no guarantee of optimal performance at the patient-level, because the metrics that define good performance are different.

A valuable literature has called out this mismatch  \citep{whoGeneratingEvidence, reyna, wiens, ehrmann, hicks, kelly, huang, misic, saha, delahuntMetrics}, often written by clinicians, who note for example, 
``The intended use of an AI-SaMD [AI-Software as Medical Device] should define, as clearly as possible, information pertaining to when, where and how it is to be used. This enables evidence generated to be evaluated in the right context for safety and performance requirements" \citep{whoGeneratingEvidence}; 
``The disconnect between the metrics for algorithm performance and the realities of a clinician's workflow and decision-making process is a fundamental but often overlooked issue" \citep{reyna}; ``Assessing clinical utility requires careful evaluation against the scenario in which the model will be used" \citep{wiens}; ``None of these measures [traditional ML metrics] ultimately reflect what is most important to patients, namely whether the use of the model results in a beneficial change in patient care" \citep{kelly}.
A comprehensive and valuable guide to evaluation metrics for ML-for-healthcare models is \citet{metricsReloaded}.

A key commonality of all these papers is: They are  concerned with how to report performance of a trained, locked model.
It is certainly vital to use clinically-tailored metrics for performance evaluation.
But at that point the impacts of earlier choices as to metrics for optimization are already baked in.

Here we address the earlier optimization stage of ML development, as a crucial but neglected opportunity to leverage clinically-tailored metrics \textit{during} optimization, \textit{before} a model is locked.
For if a model is optimized according to a Figure of Merit (FoM) that does not accurately encode the clinical requirements, the model will not attain a clinically-optimal state, because it is by definition being directed towards some other optimum. 
This is analogous to heading northwest when the actual  destination is due north: ``close'' might work if the problem is tractable enough, but for a difficult problem it carries risk that the model will fail to meet the clinical requirements.
 
This paper examines the effect of applying clinically-focused metrics, rather than validation loss, during model optimization. 

We distinguish this approach from the familiar construction of loss functions, e.g. with form $L = L_{a_1} + L_{a_2} + \alpha L_r$, where $L_{a_i}$ are variants of losses like cross-entropy (CE) and $L_r$ is a regularizer.
Though these can work well from an ML perspective, they usually must be differentiable, and this constraint can prevent tailoring them to specific, concrete clinical performance requirements.
(In the object-level vs patient-level example, if objects outnumber patients then object-level losses will also be markedly smoother than patient-level losses.)
Since many optimization tasks do not require differentiability, the choice of metric is broader.

This paper describes two controlled experiments that demonstrate a clear benefit of applying optimization metrics that are closely tailored to use case-specific demands, rather than relying on standard validation loss.
The experiments are ``real-world", drawn from deployment-focused projects.
Our goal is to clearly illustrate this opportunity to improve the clinical performance of ML models.
An example of this approach (the only one to our knowledge) can be found in \citet{buturovic}, which includes an optimization loss tailored to the particular needs of the application's regulatory pathway. 

The approach described here are readily applicable to any ML project, and are potentially valuable if the FoMs defining clinical performance are different from the loss functions that drive ML training (as is usually the case). 
The approach requires two steps:
\begin{enumerate}
\item Accurately capture the clinical performance requirements as a FoM that is a function of algorithm outputs, by combining the \{algorithm outputs $\rightarrow$ patient disposition\} function with the clinical requirements (e.g. the minimum acceptable patient-level sensitivity and specificity).
\item Inject this metric at the relevant stage to drive the model optimization or selection.
\end{enumerate}

The next two sections of the paper describe the two experiments, each in a Background-Methods-Result structure.
They are: (i) Optimizing hyperparameters using a patient-level FoM rather than an object-level FoM; and (ii) Choosing a stopping point for a DNN model using clinically-relevant metrics rather than the usual validation loss curve.

To clarify expectations, we note that the details of the particular datasets and models are not important for this paper. 
They are vehicles to illustrate the topic of metric choices for model optimization.


\section{Experiment 1: Hyperparameter optimization}
\label{sec:hyperopt}

This section describes an experiment comparing the effects of a patient-level vs object-level loss function as a driver of hyperparameter optimization. 
The details of the model are not important (they are described in 
\citet{delahuntLoaAlgorithm}). 
The important element is that an identical model architecture (trained at the object-level) has hyperparameter optimization done in two ways: driven by patient-level FoM or by object-level FoM.

\subsection{Background}
We consider an imaging device that looks at videos of fresh blood in capillaries to diagnose \textit{Loa loa} infections by detecting motion of filariae in fresh blood for the ``Test and Not Treat'' use case in Mass Drug Administration \citep{Kamgno2017}.
A clinically-important issue arose during field trials: 
If the blood in the capillary coagulates due to some delay in imaging, the filariae cannot move so the motion detection algorithm returns a potentially dangerous False Negative. For example images, see \figureref{fig:loaVideoScreenshots}. 
Therefore a module to detect coagulation was required to return a label (coagulated or not) for the patient.

\begin{figure}[!htbp]
\floatconts
  {fig:loaVideoScreenshots}
  {\caption{Frames of blood capillary videos, normal (left) and very coagulated (right). 
  The dataset contains a range of coagulations.}}
  {\includegraphics[width=0.9\linewidth]{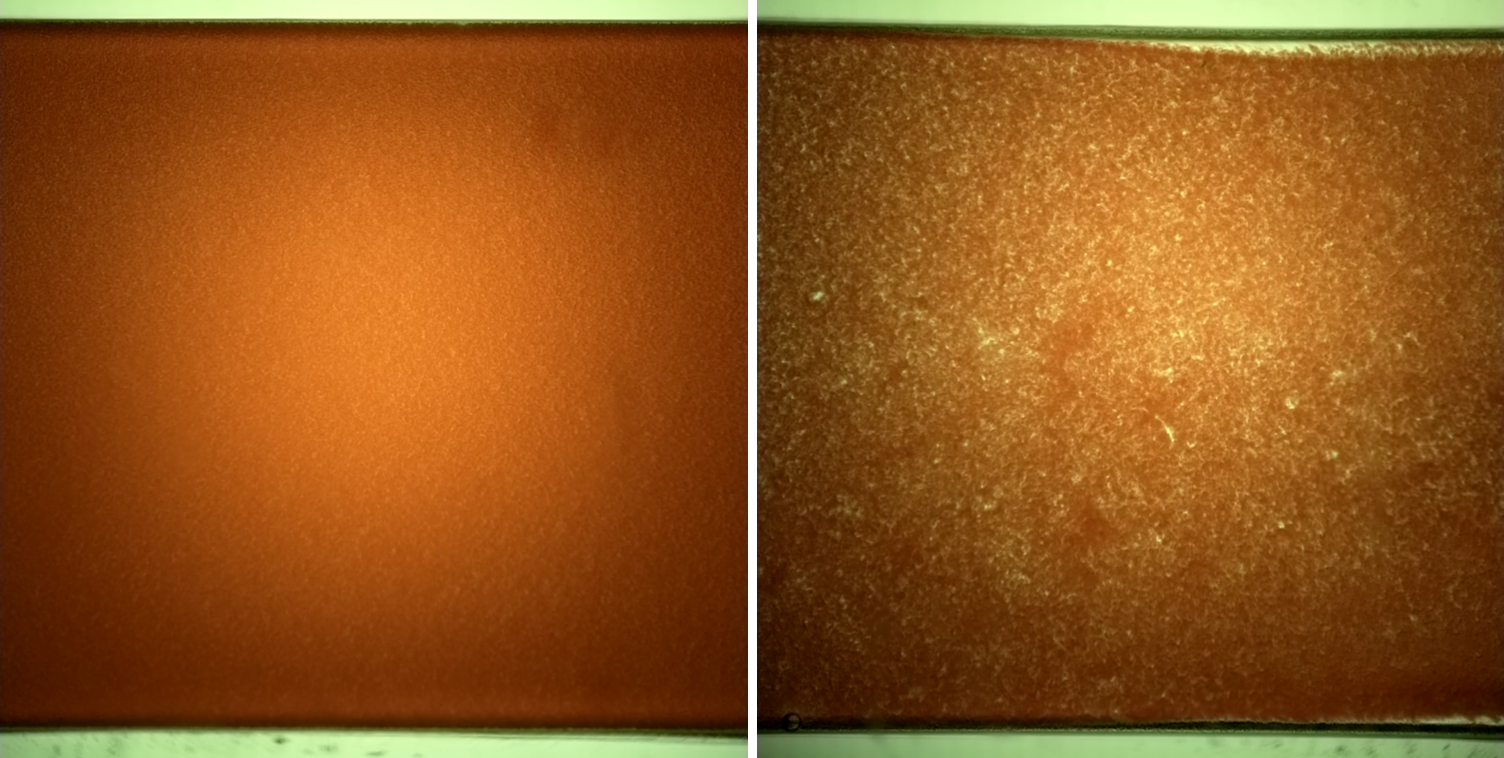}}
\end{figure}

\subsection{Methods} 
For the \textit{Loa loa} diagnosis task, the inputs are 7 videos taken along the length of the capillary.
For the coagulation task, 1 still frame is extracted from each video, so the inputs are 7 frames (7 ``objects'') per patient.

An SVM \citep{scikit-learn} acts on FFT spectrum features (chosen because coagulated blood acquires a ``checkerboard'' appearance) to give a score to each video (``video-level'', ``frame-level'', and ``object-level'' are equivalent for this task).
To assign a patient-level label, the \textit{N\textsuperscript{th}} video's score is used (\textit{N} is a hyperparameter).
So the SVM is trained at the object-level to classify objects (videos) as coagulated or not. 
But the required clinical output is a patient-level classification, which is a non-linear function of the object scores.

Model hyperparameters include SVM parameters, feature selection parameters, and a few others. 
For optimization we used the hyperopt library \citep{hyperoptLibrary}, which is guided by the Tree of Parzen Estimators method \citep{bergstra_hyperparameters}. Differentiability is not required.

For this experiment, everything in the set-up is kept fixed except for the choice of loss function to drive the hyperopt optimization: in one case AUC over videos is used, matching the object-level of the model's training\footnote{The SVM's hinge loss is not visible. Our manual hinge loss gave results similar to object-level AUC.}; in the other, AUC over patients is used, matching the clinically-relevant output. 
(The actual loss used by hyperopt is 1 - AUC.)


Results are reported for the validation sets in a 5-fold cross-validation.
The different fold scores are combined and made comparable using a novel (to our knowledge) z-mapping technique.
The method is described in \appendixref{apd:zScalingKFold}.

\subsection{Results}
The crucial finding is that patient-level and object-level optima are not correlated, and optimizing the object-level metric, though perfectly sensible from a ML perspective (since the model is trained at object-level) leads to inferior patient-level performance. 
The following figures illustrate this disconnect between object-level and patient-level losses:

\begin{enumerate}
\item \figureref{fig:rocsPatientAndVideo} shows the ROC curves for the best iteration in the video-level run (on left) and for the best iteration in the patient-level run (on right).
The video-level ROCs are very similar, but the patient-level ROC is much better when patient-level metrics drive the optimization.
\item \figureref{fig:sortedVsUnsorted} \textbf{A} shows the patient-level and object-level AUC values for each iteration of the hyperopt run driven by patient-level AUC, with iteration order sorted by increasing patient-level AUC.
Note that the object-level AUCs are simply a cloud: optimizing patient-level AUC does not optimize object-level AUC.
\item \figureref{fig:sortedVsUnsorted} \textbf{B} shows the equivalent plot for the hyperopt run driven by object-level AUC, sorted by increasing object-level AUC.
As above (but reversed), optimizing object-level AUC does not optimize patient-level AUC.
\item \figureref{fig:sortedVsUnsorted} \textbf{C} shows a scatterplot of patient-level vs. object-level AUCs per iteration (from the object-level run). 
Note the lack of correlation between object- and patient-level results.
\end{enumerate}
 
Thus, if the goal is to optimize clinical performance it is best to drive hyperparameter optimization not by the model's loss function, but by a loss function that more closely aligns with clinical requirements.
 
\begin{figure}[!hbtp]
\floatconts
  {fig:rocsPatientAndVideo}
  {\caption{Best ROC curves for patient-level (red) and video-level (black).
  Per subplot, the two curves are from the best hyperopt iteration, as driven by: \textbf{(left)}  object-level AUC, \textbf{(right)} patient-level AUC.
  Patient-level optimization gives much better patient-level performance.}}
  {\includegraphics[width=0.88\linewidth]{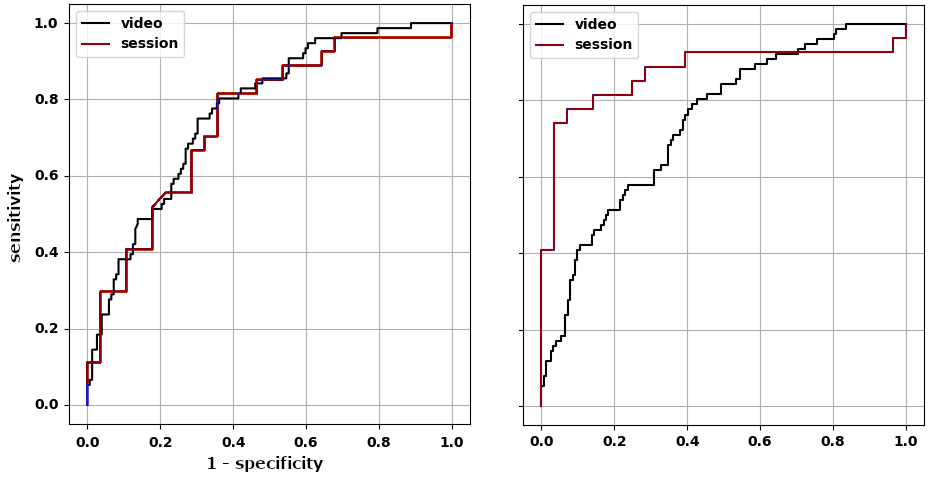}}
\end{figure}

\begin{figure*}[!hbtp]
\floatconts
  {fig:sortedVsUnsorted}
  {\caption{Patient-level and video-level AUC values for different iterations of hyperopt. x-axis = sorted iteration index, y-axis = AUC value. Patient-level AUCs in green, video-level AUCs in blue. \textbf{A}: Optimizer driven by patient-level AUC, iterations sorted by patient-level AUC values. \textbf{B}: optimizer driven by video-level AUC, iterations sorted by video-level AUC values. \textbf{C}: Scatterplot of patient-level vs video-level AUCs for data in subplot B. Note the lack of correlation between the two performance metrics.}}
  {\includegraphics[width=1\linewidth]{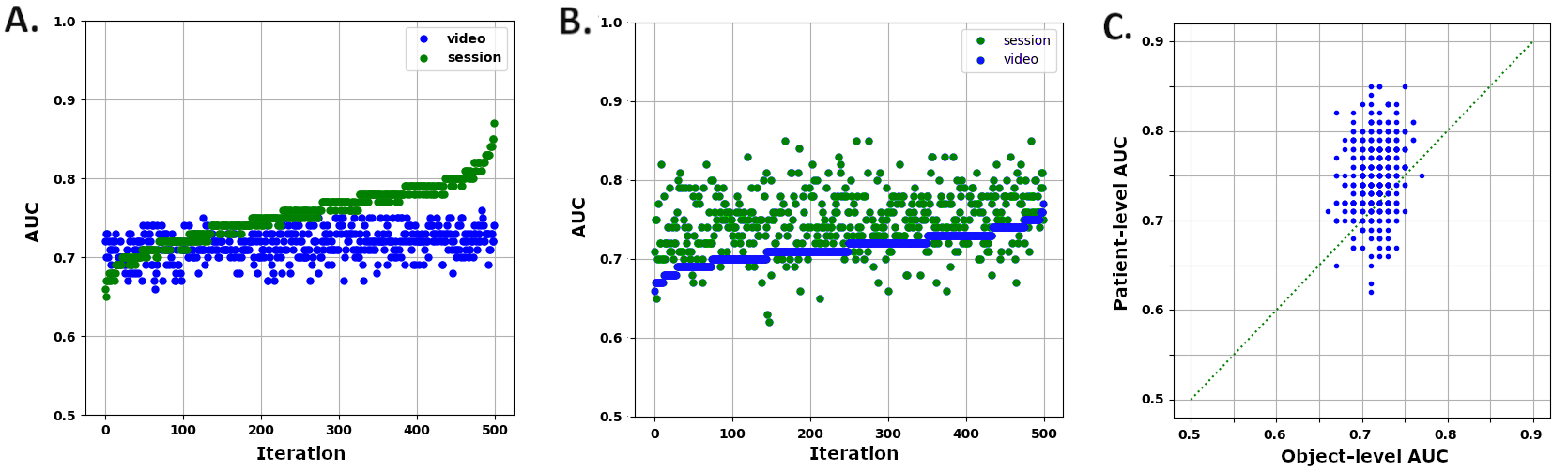}}
\end{figure*}


\section{Experiment 2: Stopping point optimization}
\label{sec:stopping} 

\subsection{Background}
DNNs are trained over several epochs, and an important optimization task is to choose the best epoch's model for testing or deployment. 
The usual method is to consult the loss curves for train and validation sets, where the same loss function is applied to both train and val sets.
Then the final epoch (i.e. model) is chosen based on the validation loss curve, e.g. where it starts to increase, implying that overfitting has begun.
This is an effective method to choose the best model in the narrow sense of performance as measured by the training loss function. 

But clinical performance requirements are not, in general, captured by the training objective.
Although the loss is often structured to generally reflect the medical context, it is also often shaped by reasons of convention, differentiability, or ease of implementation (e.g. the existence of built-in functions).

Therefore this experiment assesses whether the validation loss curve is in fact an optimal guide to choosing the best epoch in terms of performance on the clinical task. 
We train a DNN for 40 epochs and then plot, for each epoch,  multiple FoMs: not just the validation loss but also other metrics that more tightly reflect the clinical use case.
We also plot per-epoch histograms of the validation exam scores to assess the stopping points selected by each FoM.

As before, the details of the model are not important (they are described in 
\citet{mehanianObus}).
The important point is that an identical model has its stopping point optimized in one of two ways: with the usual validation loss curve, or with clinically-relevant metrics along with per-epoch assessment plots such as score histograms.

We consider the case of a DNN trained to distinguish twins vs singleton fetuses using blind (unguided) ultrasound sweep videos. 
This is clinically important because twin pregnancies tend to be higher risk and should be identified. 
The blind sweeps with algorithms enable this diagnostic in low resource settings where trained sonographers are not available.

\subsection{Methods}
 
The data consist of blind sweep ultrasound exams collected in Zambia and the USA by a team led by U. of North Carolina's Global Women's Health group \citep{pokaprakarn_blindSweep}.
Twins are the ``positive'' class, singletons are the control.

A DNN architecture delivers an exam-level score. It is trained in Pytorch \citep{pytorch2019} and Pytorch Lightning \citep{pytorchLightning}  with balanced cross-entropy loss. 
The Pytorch library, as typical for ML frameworks, calculates the same CE loss on the validation set, and produces a validation set loss curve. We plot 1 -  validation loss.

We also plot alternate metrics tailored to the clinical goal.
To generate these metrics requires some extra effort.
First, the exam scores must be saved at each epoch, either via a custom callback function during training, or by running each epoch's model on the validation set as inference. 
Then custom code is needed to calculate metrics and generate plots. 
This is a cost of extending optimization beyond built-in ML functions.
The following FoMs are plotted:
\begin{enumerate} 
    \item Validation loss returned by PyTorch (here, balanced CE).
    \item Standard area under the ROC curve (AUC). 
    \item 90\% ``sliver'' AUC. 
    The sliver AUC considers only the subset of the AUC where specificity is $>90\%$, i.e. the area under the ROC in the leftmost $1 / 10^{th}$ of the normal ROC. 
    This FoM is valuable because the minimum acceptable clinical specificity (for this project) was 90\%, so the entire righthand region of the ROC is irrelevant clinically. 
    
    The $n\%$ sliver AUC can be calculated by summing trapezoids over the operating points with False Positive Rate between 0.0 and $(1 - n / 100)$, normalized by $(1 - n / 100)$.    
    For an example of sliver AUC, see \figureref{fig:sliverAucs}. 
    Code is provided in \appendixref{apd:codeFragments}.  
    \item Sensitivity at 90\% specificity. 
    If the clinically-acceptable specificity is known (here 90\%), the corresponding sensitivity is a scalar that directly measures the model's potential performance at a clinically-relevant operating point. See \citet{whoTbScreening} for a (non-ML) example from tuberculosis, where sensitivity is set to 90\% and the corresponding specificity is evaluated.
    \item Fisher distance, defined for two distributions by 
    \[\frac{| \mu_1 - \mu_2|}{\sigma_1 + \sigma_2} \text{ where } \mu_i,~ \sigma_i = \text{ means and standard}\]
    deviations.
    We use right and lefthand std devs which are useful for asymmetric distributions (code is given in \appendixref{apd:codeFragments}).
\end{enumerate}

We also plot, per epoch, some histograms and scatterplots, because these are useful for qualitative assessment of what the models are doing relative to the FoMs given above.
\begin{enumerate}
    \item Scatterplots and histograms of scores, separated by class. 
    \item Plots of sensitivity and specificity curves vs threshold on scores.
    \item Scatterplots of scores vs gestational age (GA).
\end{enumerate}

\begin{figure}[!hbtp]
\floatconts
  {fig:sliverAucs}
  {\caption{Sliver AUC example: 
  The blue ROC has lower overall AUC than the red (0.9 vs 0.96), but it is better at high specificities and thus has a higher 90\% sliver AUC  (in the grey column, 0.82 vs 0.77).}}
  {\includegraphics[width=0.7\linewidth]{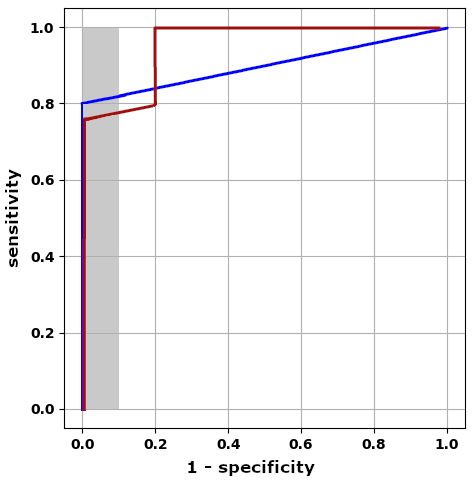}}
\end{figure} 

\subsection{Results}

The crucial finding is that, when these various FoMs are calculated at each epoch on the validation set, the validation loss curve suggests a substantially different ``best'' stopping point than do the clinically-tailored FoMs.
In addition, the per-epoch scatterplots and sensitivity-specificity indicate that the stopping point suggested by the training loss is too early and will likely give inferior results in the clinical task.

A 5-fold split was used for cross-validation purposes, so 5 models were trained. 
Results are given below for one of the 5 folds - all folds showed similar behavior.

The disagreement between ``best'' stopping points is seen in the per-epoch time-series of PyTorch's val loss and the alternate FoMs, shown in \figureref{fig:fomTimeseries} (the negative of val loss is shown, so that for all FoMs higher is better).
The val loss maxes out at epoch 22, while the the clinically-tailored FoMs keep steadily increasing up to epoch 39.
Specifically:
\begin{enumerate}
\item The validation loss (using the training loss function) maxes out at epoch 22, then decreases in a distinct though noisy way.
\item The standard AUC maxes out at around epoch 20, and shows no further change after this. 
That is, standard AUC cannot distinguish between the models of epochs 20 - 39.
\item The 90\% sliver AUC shows steady increase up to epoch 39, i.e. it argues for later epochs.
\item The ``sensitivity at 90\% specificity'' FoM increases until epoch 29, then mostly stays at this maximum. 
So it argues for $\ge29$ epochs.
\item The Fisher Distance matches the behavior of the sliver AUC, steadily increasing to a maximum at epoch 35.
\end{enumerate}

\begin{figure*}[!htbp]
\floatconts
  {fig:fomTimeseries}
  {\caption{Figure of Merit time-series over training epochs. 
  x-axis: epoch indices, y-axis: FoM value. 
  \textbf{L - R:} (1) -1 * Standard validation loss; (2) AUC; (3) 90\% Sliver AUC; (4) Sensitivity at 90\% specificity; (5) Fisher distance. 
  x-axis: epoch. y-axis: value (higher is better). Highest scores marked in red. Clinically-relevant FoMs suggest much later stopping points than that of validation loss.}}
  {\includegraphics[width=1\linewidth]{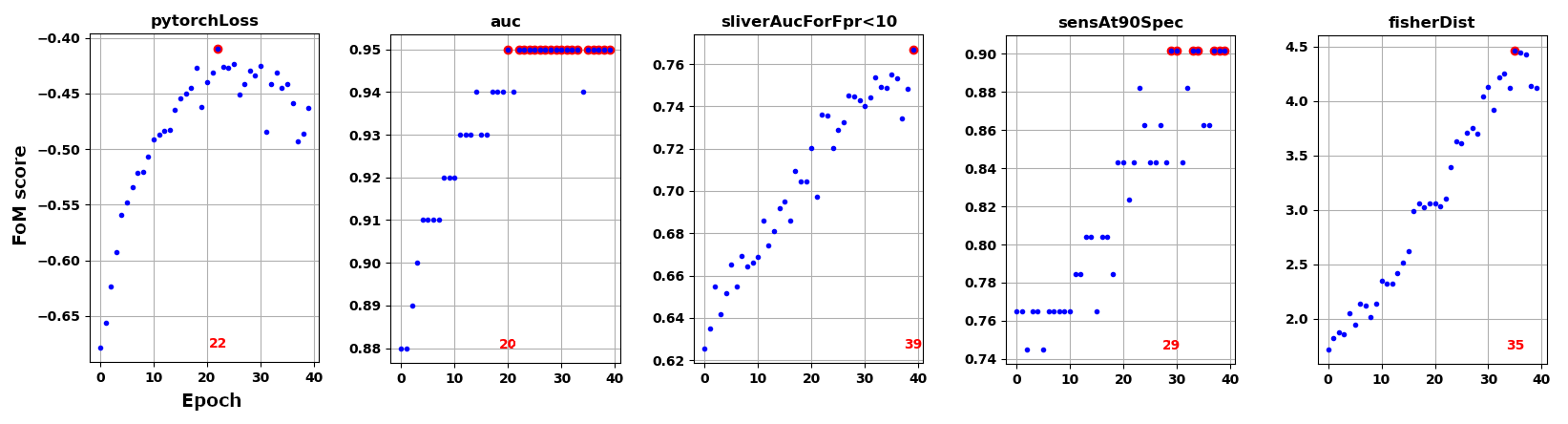}}
\end{figure*}

The question now arises: Which stopping point is in fact the best?
The separation behavior of the per-epoch models, for epochs \{0, 3, 12, 22, 29, 35, 39\}, are shown in the scatterplots and histograms in \figureref{fig:scatterplotsAndHistograms}.
Epoch 22, despite having the best validation loss, has poor separation compared to epochs 29 - 39.
 
The later epochs also show greater stability around the clinically-relevant operating point, seen in the per-epoch sensitivity and specificity plots of \figureref{fig:sensSpecPlots}.
Stability is indicated by shallower slopes in the two curves at the required operating point ($\ge 90\%$ specificity), since this reflects greater robustness to changes in operating point.
Epoch 35 is strongest in this regard.

\begin{figure*}[!htbp]
\floatconts
  {fig:scatterplotsAndHistograms}
  {\caption{Exam score distributions per epoch: \textbf{Top:}Score scatterplots: Singletons in green, twins in red. y-axis = model score.
  \textbf{Bottom:} Histograms of scores: Singletons in green, twins in red.
  Class separation continues to improve after epoch 22 (which has best validation loss), indicating that the model is still usefully learning.}}
  {\includegraphics[width=1\linewidth]{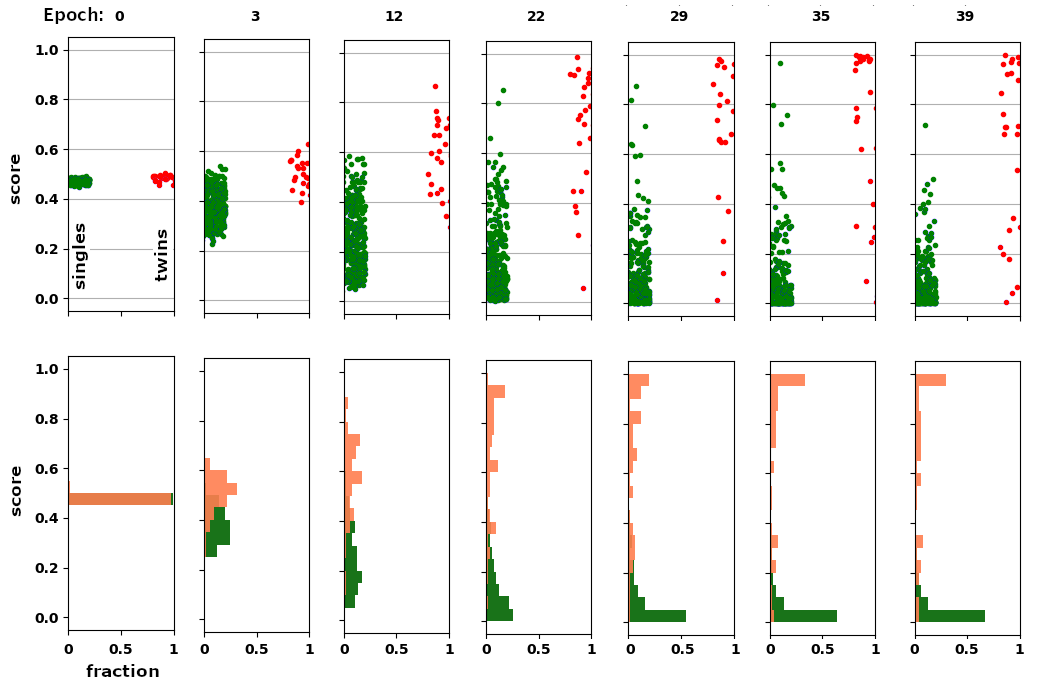}}
\end{figure*}
 
\begin{figure*}[!htbp]
\floatconts
  {fig:sensSpecPlots}
  {\caption{Sensitivity (red) and specificity (green) vs threshold, per epoch. x-axis: threshold. y-axis: value (of sens or spec) 
  Shallower curves around the clinically-relevant operating point ($\ge90\%$ specificity, marked by short gray lines) indicate more stability.}}
  {\includegraphics[width=1\linewidth]{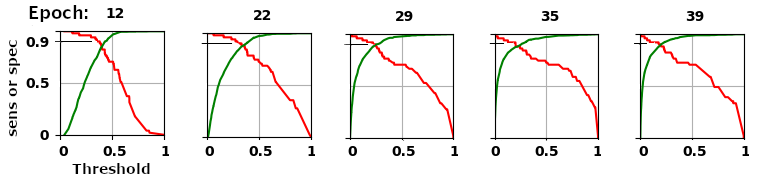}}
\end{figure*}

An interesting effect specific to the particular dataset/use-case is seen in \figureref{fig:scatterScoreVsGa}, which shows per-epoch scatterplots of scores vs GA. 
Twin fetuses with low GA, e.g. $<80$ days, are much harder to classify for biological reasons, and all epochs give lower scores to low GA cases.
However, the model at epoch 35 ``sacrifices'' these cases, giving them very low scores, in order to further reduce singletons' scores, ensuring better overall separation.
By contrast, epoch 22 has higher scores for low GA cases, but at the cost of spread-out, higher scores for singletons. 
Thus, epochs $\ge35$ separate the two classes better, except for the ``lost cause'' low-GA cases.  
This effect is germane to the clinical use case, so the ability to notice it enables better optimization for the clinic.

In summary: The standard validation loss curve suggests stopping relatively early, but the clinically-tailored FoMs, backed up by the behaviors in the per-epoch assessment plots, indicate that training much longer is better for the clinical task.
This crucial information is not revealed by the validation loss curve.

\begin{figure*}[!htbp]
\floatconts
  {fig:scatterScoreVsGa}
  {\caption{Exam scores vs GA, per epoch: x-axis = GA. y-axis = score. Red = Twin, green = singletons. Low GA cases are always difficult. Later epochs (e.g. 35 vs. 22) ``sacrifice'' them to force singleton scores closer to 0 and twin scores closer to 1, giving overall better separation.}}
  {\includegraphics[width=1\linewidth]{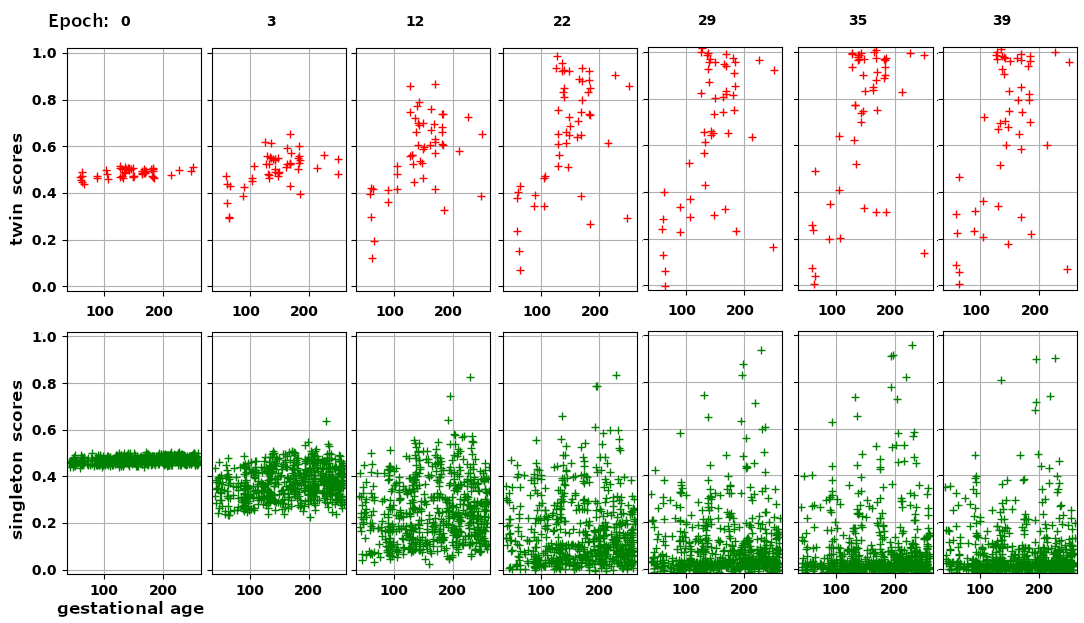}}
\end{figure*}


\section{Discussion}
\label{sec:discussion}
A key task in ML is to optimize of models at various stages, e.g. by choosing hyperparameters or picking a stopping point.  
A standard ML approach is to use the training loss function on a validation set (validation loss) to guide these optimizations, because this loss meets the differentiability requirements of backprop and gradient descent, and also perhaps because of habit, convention, and ease of use (e.g. with built-in functions and libraries).

However, ML for healthcare is a distinct field from traditional ML, and it has a distinct goal: The model performance must meet rigorous clinical specifications.
These demands can be captured more precisely by metrics customized to the clinical needs than by standard ML objective functions.
Since many optimization tasks do not require differentiability, they admit of a wider range of metric options.

In this paper we gave two examples of how metrics tailored to be clinically-relevant provide superior model optimization, in the sense of giving better performance on the clinical task (they give worse performance in the sense of an ML-centric task definition). 
This result is not surprising: To excel at a task, it makes sense to tailor the optimization to that task. 
For ML applied to health care, this requires moving beyond the metrics and conventions inherited from traditional ML.
 
Implementation of this approach is necessarily a bespoke process, because the integration details are highly task-specific, depending on both the particular clinical use case and the chosen ML framework. 
Thus, a close understanding and devout centering the clinical use case is the first step, followed by identifying or deriving usable Figures of Merit that encode the specifics of the use case.
This involves careful definition of the metrics required for clinical deployment, and also how the outputs of a model feed into a clinically-relevant disposition (e.g. the object-level to patient-level transform). 

Sometimes an existing metric will work. For example, \textit{n}\% sliver AUC and sensitivity at \textit{n}\% specificity may apply since many diagnostic use cases have well-defined specificity requirements. Absent a pre-existing metric, the solution is careful understanding of the specific clinical context and some math.
An example of a manual derivation (for malaria) is in \citet{delahuntMetrics}.
See also examples in \citet{ehrmann, misic, saha}.

Consultation with non-ML literature and domain experts, before any ML coding, can identify relevant quantitative clinical requirements.
Examples: 

(i) Our treatment of coagulation was closely grounded in the details of that specific use case, as described by field teams; 

(ii) The ultrasound metrics were constructed to capture the sensitivity / specificity targets given for the OB ultrasound project.
These targets were based on consultation with clinicians, landscaping of the use case (e.g. strategy documents from groups like the Gates Foundation), and the stage of our work in the dev-to-regulatory pipeline;

(iii) \citet{buturovic} incorporates a specific regulatory constraint into their optimizer metric.

\smallskip
We deliberately do not advocate patient-level training losses in this paper, because an object-based training loss plus a separate object-to-patient linkage is often more effective than (and not equivalent to) a direct patient-level loss, for at least three reasons: 

(i) In medical datasets the patient count is often low, and there are also typically many fewer patients than objects. In experiment 1 it is 7-to-1, but for use cases involving parasites or abnormal cells it can be thousands-to-1. So a patient-based loss can be too granular due to the relatively few individual losses in the sum, while an object-based loss has smooth behavior due to the high number of objects. 

(ii) Objects typically present a fundamentally easier target for the ML model. For example, CNNs are beautifully suited to identifying object thumbnails, but directly identifying an ``infected patient" is harder to frame as an ML task.

(iii) Some medical tasks involve identification of objects by the clinician. In these use cases, identifying and showing those objects to the clinician both serves as a valuable interpretive trust-building tool, and opens valuable options for merging into the clinical workflow (e.g. as clinician support vs automated decision). 
       
\smallskip
The use of clinically-relevant metrics for model optimization steps entails some extra effort, to define appropriate metrics and code them into the pipeline.
But in return it can yield models that better meet the central goal of ML for healthcare: Optimal performance in the clinic. 

\acks{
Like all ML researchers, we owe everything to those who create the datasets: 

Our thanks to Dr. Joseph Kamgno and his team at the Center for Research on Filariasis and other Tropical Diseases (CRFilMT) in Cameroon for collecting the \textit{Loa loa} dataset, to all the generous participants, and to Anne-Laure Nye for annotations.

Our thanks to Dr. Jeffrey Stringer, the Global Women's Health group at U. of North Carolina School of Medicine, and their Zambian collaborators for collecting and curating the FAMLI dataset, and to all the mothers-to-be who generously participated. \\
 
\noindent This work was funded by Global Health Labs, Inc. (\url{www.ghlabs.org}).
}

\bibliography{beyondValidationBibliography}

\onecolumn

\appendix

\section{Scaling method to combine k-folds}\label{apd:zScalingKFold}

A \textit{K}-fold split setup yields \textit{K} distinct models and  associated scores on their respective validation sets. 
Each sample has exactly one output score when treated as a validation sample, as calculated by one of the \textit{K} models. 
Performance of the \textit{K} models on their validation sets offers valuable insight into model variability.
But these per-model results can be highly volatile for medical datasets that have small patient counts.

In such cases, it can be desirable to combine all the samples into one set, to give a larger validation set on which to evaluate model performance.
The catch is that the different models are calibrated differently, i.e. they have different score ranges so that a given operating point will correspond to different thresholds in each split, as seen in the left subplot of \figureref{fig:zScaleExample}.
Therefore we offer here a technique to combine the  model scores on their validation sets into a single consistent set of scores for which a single threshold can be used.

We assume two classes, one of which is the ``control'' class.
For example, blood samples that are uncoagulated (the control) or coagulated, with class labels 0 and 1.
Let samples be ${s_{i, k}}^c$, where $c \in \{0,1\}$ is the class, $i$ is the sample index, and $k$ is the split.
The technique basically consists of mapping, for each split, the distribution of its control sample validation scores to a common z-scale and common median.
The coagulated sample scores also get mapped according to this transform, i.e. they ``go along for the ride''.

\begin{enumerate}
\item For each fold \textit{k}, calculate the median and the two-sided standard deviations of the control samples, $m_k = \text{median}({s_i}^0$) and similarly $\sigma_{r, k}, \sigma_{l, k}$. See code in \appendixref{apd:codeFragments}.
\item Choose a target median and right-hand standard deviation $m_t$ and $\sigma_{r,t}$, e.g. $m_t = 0.3, \sigma_{r,t} = 0.2$.
\item For each \textit{k}, scale all the sample scores in the \textit{k\textsuperscript{th}} validation set to a new normalized score $n_i$:\\
$n_i = \sigma_{r, t}({s_{k, i}}^c - m_k) / \sigma_{r, k}$ 
(if $s_{k, i} > m_k$; similarly use $\sigma_{l,t}$ if $s_{k, i} < m_k$). 
\end{enumerate}
The $\{n_i\}$ are all directly comparable, in the following sense:
if $s_{i, k_1}$ is $x$ std devs above $m_{k_1}$ and $s_{j, k_2}$ is $x$ std devs above $m_{k_2}$, then $n_i = n_j$, so an $x$ std dev threshold treats them both the same, whether it is in fold $k_1, k_2$, or in the common scale.
See \figureref{fig:zScaleExample} for an example.
Code is provided in \appendixref{apd:codeFragments}.

This method works best when scores are not already pushed to the rails (if scores are clustered at 0 and 1, the method has little impact vs. simply combining the various folds' scores as-is). 

\begin{figure*}[!htbp]
\floatconts
  {fig:zScaleExample}
  {\caption{Effect of z-scale alignment. y-axis = scores. Green = controls, red = coagulated. \textbf{Left:}  Raw output scores per fold: Each fold's model requires a different threshold for a given specificity. \textbf{Right:} mapped scores per fold: One threshold gives the same specificity for all folds.}}
  {\includegraphics[width=0.8\linewidth]{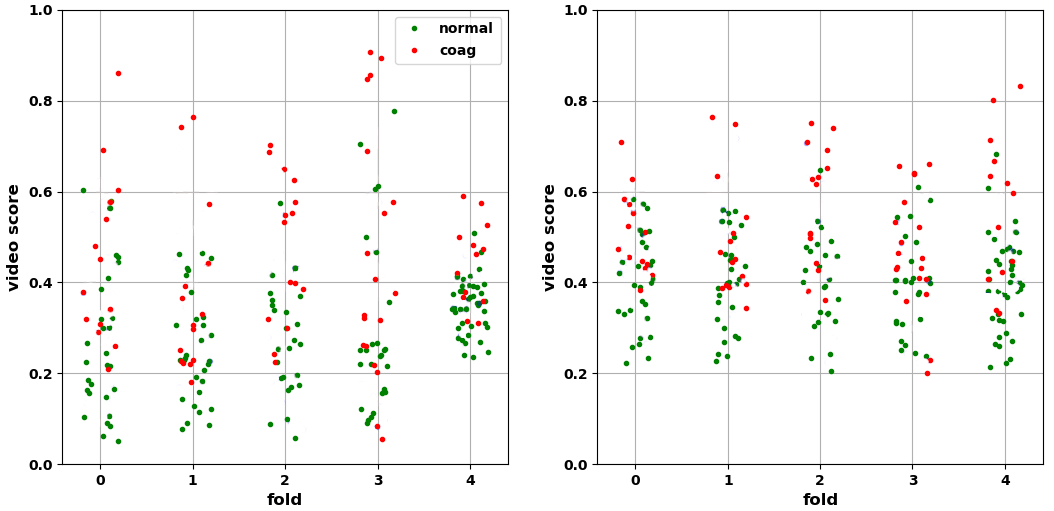}}
\end{figure*}

\section{Python function defs}\label{apd:codeFragments}

\subsection{2-sided standard deviations}

Asymmetrical distributions are imperfectly described by the standard deviation with gaussian assumption.
A quick way to more precision is to calculate two standard deviations, right- and lefthanded, using just the points to the right (or left) of the median.  \\

{\fontfamily{cmss}\selectfont

\noindent import numpy as  \\
\noindent def calculateTwoSidedStdDev\_fn(x, middleVal=None)  \\
\indent"""\\
\indent Calculate two standard deviations, one for the left and one for the right, by flipping samples across \\
\indent the median (not across the mean, on the grounds that in asymmetrical distributions the median  \\
\indent is likely closer to the peak value). \\ \\
\indent Parameters \\
\indent ---------- \\
\indent x : list-like or np.array vector \\
\indent middleVal: float or int. If you don't want to use the median. Default = None, ie use median. \\ \\
\indent Returns \\
\indent ------- \\
\indent stdDevRight : scalar float \\
\indent stdDevLeft : scalar float \\
\indent """ \\ \\        
\indent if len(x) $<=$ 1: \\
\indent \indent stdDevRight = -1 \\
\indent \indent stdDevLeft = -1 \\
\indent else: \\
\indent \indent if middleVal != None: \\
\indent \indent \indent m = middleVal \\
\indent \indent else: \\
\indent \indent \indent m = np.median(x) \\
\indent \indent x = x - m \\
\indent\indent xH = x[x $>=$ 0]  \# top half, shifted to a nominal 0 center \\
\indent\indent xL = x[x $<=$ 0]  \# bottom half \\ \\
\indent \indent stdDevRight = np.std(np.concatenate((-xH, xH))) \\
\indent \indent stdDevLeft = np.std(np.concatenate((xL, -xL))) \\ \\
\indent return stdDevRight, stdDevLeft   \\  \\    
\noindent \# End of calculateTwoSidedStdDev\_fn \\    
}

\subsection{z-scale alignment}
The first def below extracts the parameters of the distributions in each split.
The second def applies these parameters to z-map the scores of each split.   

{\fontfamily{cmss}\selectfont

\noindent import numpy as np  \\ \\
\noindent def standardizeSvmScoresForKArrayGivenParams\_fn(x, p):  \\
\indent """  \\
\indent Given an array of model output scores from k splits adjust them using parameters   \\
\indent pre-calculated by 'calculateStandardizationParamsForKFoldScores\_fn' (above).  \\
\indent We apply vector operations to, for each split's score,  \\
\indent \indent 1. Subtract that split's median to center the scores at (hypothetical) zeros   \\
\indent \indent 2. Scale each values by its split's stdDevs (RH and LH) to roughly match the canonical std dev  \\
\indent \indent 3. Shift all the values to the canonical median.  \\
\indent NOTE: Function exits if there is a zero in splitRhStdDevs or splitLhStdDevs. This should not occur,   \\
\indent because the training samples that generated the std devs would have to have identical scores.  \\
\indent Parameters  \\
\indent ----------  \\
\indent x : list of floats, len = number of splits, ie number of models \\
\indent p : dict with keys 'canonicalMedian', 'canonicalStdDev', 'splitMedians',  \\
\indent \indent 'splitRhStdDevs', 'splitLhStdDevs'  All entries except 'canonicalMedian' and  \\
\indent \indent 'canonicalStdDev' will be vectors if we are adjusting all splits at once. \\ \\
\indent Returns \\
\indent ------- \\
\indent adjX : list of floats, same length as argin 'x'. \\ \\
\indent """ \\
\indent rhStd = np.array(p['splitRhStdDevs']) \\
\indent lhStd = np.array(p['splitLhStdDevs']) \\
\indent canonStd = np.array(p['canonicalStdDev']) \\
\indent canonMedian = np.array(p['canonicalMedian']) \\
\indent splitMedians = np.array(p['splitMedians']) \\ \\    
\indent if np.sum(rhStd == 0) $>$ 0 or np.sum(lhStd == 0) $>$ 0: \# should not happen \\ 
\indent \indent print('splitRhStdDevs or splitLhStdDevs contains a 0 value. Returning input scores.') \\
\indent \indent adjX = x \\
\indent else: \\
\indent \indent t0 = x - splitMedians  \# subtract each split's median \\
\indent \indent \# Apply column vector operations: \\
\indent \indent for k in range(len(splitMedians)): \# Process each column in turn: \\
\indent \indent \indent tk = t0[:, k] \\
\indent \indent \indent if np.sum(tk $>$ 0) $>$ 0: \\
\indent \indent \indent \indent tk[tk $>$ 0] = tk[tk $>$ 0] * canonStd / rhStd[k]  \# to right of median \\    
\indent \indent \indent if np.sum(tk $<$ 0) $>$ 0: \\
\indent \indent \indent \indent tk[tk $<$ 0] = tk[tk $<$ 0] * canonStd / lhStd[k]   \# to left of median \\ 
\indent \indent \indent t0[:, k] = tk \\             
\indent \indent adjX = t0 + canonMedian * np.ones(x.shape) \\ \\
\indent return adjX \\ \\
\noindent \# End of standardizeSvmScoresGivenParams\_fn\\ 
\noindent \#-----------------------------------------  \\  \\
%
%
\noindent def standardizeKFoldScoresForVectorGivenParams\_fn(x, split, p):  \\
\indent """  \\
\indent Given a vector of scores from a model, along with a vector of their splits, standardize the   \\
\indent scores of each fold using the dictionary of parameters.   \\ 
\indent The argins will likely be the scores on k test sets, concatenated into a vector, where the  \\
\indent scores came from k models built on a k-fold split.  \\  \\
\indent Parameters  \\
\indent ----------  \\
\indent x : list-like of floats (scores)   \\
\indent split : list-like of ints (fold indices). same len as 'scores'  \\
\indent p : dict of params (see unpacking in first few lines)  \\  \\
\indent Returns  \\
\indent -------  \\
\indent adjScores : list-like of floats. same len as 'scores'   \\ 
\indent """  \\ \\
\indent foldVals = np.unique(split)  \# hopefully 0,1,2, etc. But does not need to be.   \\
\indent canonicalMedian = p['canonicalMedian'] \\
\indent canonicalStdDev = p['canonicalStdDev']  \\
\indent splitMedians = p['splitMedians']  \\
\indent splitRhStdDevs = p['splitRhStdDevs']  \\
\indent splitLhStdDevs = p['splitLhStdDevs']  \\  \\
\indent adjX = x.copy()  \\ 
\indent for k in range(len(foldVals)):  \\ 
\indent \indent \# Adjust all scores in this fold:  \\ 
\indent \indent inds = split == foldVals[k]  \\  
\indent \indent t = x[inds]  \\  
\indent \indent m = splitMedians[k]  \\  
\indent \indent r = splitRhStdDevs[k]  \\  
\indent \indent l = splitLhStdDevs[k]  \\         
\indent \indent t0 = t - m   \\
\indent \indent if r $>$ 0:  \# in case r == 0, don't divide (a catch)  \\
\indent \indent \indent t0[t0 $>$ 0] = t0[t0 $>$> 0] * p['canonicalStdDev'] / r  \# to the right of median   \\
\indent \indent else:  \\
\indent \indent \indent pass  \\
\indent \indent \indent print('rh stdDev = 0 on ' + str(np.sum(t0 $>$ 0)) + ' cases > median.')  \\
\indent \indent if l $>$ 0:  \# ditto  \\
\indent \indent \indent t0[t0 $<$ 0] = t0[t0 $<$ 0] * p['canonicalStdDev'] / l  \# to the left of median   \\
\indent \indent else:  \\
\indent \indent \indent pass  \\
\indent \indent \indent print('lh stdDev = 0 on ' + str(np.sum(t0 $>$ 0)) + ' cases $<$ median.')  \\
\indent \indent adjX[inds] = t0 + p['canonicalMedian']  \\  \\        
\indent \# optional: could clip adjX at 0 and 1.  \\
\indent return adjX   \\  \\    
\noindent \# End of standardizeKFoldScoresForVectorGivenParams\_fn \\ 
}
\subsection{n\% sliver AUC} 

{\fontfamily{cmss}\selectfont

\noindent import numpy as np \\
\noindent from sklearn.metrics import roc\_curve \\
\noindent def calculateSliverAuc\_fn(y, yHat, targetSpec): \\
\indent """ \\
\indent Given scores, binary labels, and a minimum specificity: calculate the normalized AUC \\ 
\indent within the leftmost sliver of an ROC curve. \\ \\
\indent Parameters \\
\indent ----------  \\
\indent y : list-like of ints (0s and 1s)    \\
\indent yHat : list-like of floats (scores). same len as 'y'  \\ 
\indent targetSpec : int (1 to 99)\\
\indent Returns  \\
\indent -------  \\
\indent sliverAuc : float \\ 
\indent """  \\ \\
\indent maxFprForAucLoss = (100 - targetSpec) / 100 \\
\indent [fpr, tpr, op] = roc\_curve(y, yHat, pos\_label=1)  \\  
\indent inds = np.where(fpr $<=$ maxFprForAucLoss)[0] \\
\indent f = list(fpr[inds]) + [maxFprForAucLoss]  \# postpend an endpoint fpr  \\
\indent t = list(tpr[inds]) + [tpr[inds[-1]]]  \# postpend the corresponding tpr value \\
\indent sessionSliverAuc = 0 \\
\indent for i in range(len(f) - 1): \\
\indent \indent sliverAuc += (f[i+1] - f[i]) * (0.5 * (t[i+1] + t[i])) \# Trapezoid rule \\
\indent sliverAuc = sliverAuc / maxFprForAucLoss \# normalize \\ \\
\indent return sliverAuc \\ \\
\noindent \# End of calculateSliverAuc\_fn
}

\end{document}